\documentclass[twoside,11pt]{article}

\usepackage{jmlr2e}

\usepackage{times}
\usepackage{graphicx} 
\usepackage{subfigure} 

\usepackage{natbib}

\usepackage{algorithm}
\usepackage{algorithmic}
\usepackage{color}

\usepackage{tikz}

\usepackage{amsmath}
\usepackage{amssymb}
\usepackage{epstopdf}

\newtheorem{thm}{\rm\bf Theorem}
\newtheorem{lem}[thm]{Lemma}
\newtheorem{prop}[thm]{Proposition}
\newtheorem{coro}[thm]{Corollary}

\newcommand{\dsum}{\displaystyle\sum}
\newcommand{\mbx}{\mathbf{x}}
\newcommand{\mbw}{\mathbf{w}}
\newcommand{\mbz}{\mathbf{z}}

\newcommand{\mba}{\mathbf{a}}
\newcommand{\mbb}{\mathbf{b}}

\newcommand{\dmin}{\displaystyle\min}
\newcommand{\dmax}{\displaystyle\max}

\firstpageno{1}

\begin{document}

 
\title{On the Compressive Power of  Deep Rectifier Networks \\ for  High Resolution Representation of Class Boundaries}

\author{\name Senjian An \email senjian.an@uwa.edu.au \\
       \addr School of Computer Science and Software Engineering\\
       The University of Western Australia\\
       \AND
       \name Mohammed Bennamoun  \email mohammed.bennamoun@uwa.edu.au \\      
\addr School of Computer Science and Software Engineering\\
      The University of Western Australia\\
       \AND
       \name Farid Boussaid \email farid.boussaid@uwa.edu.au \\       
\addr School of Electrical, Electronic and Computer Engineering\\
       The University of Western Australia\\
              }

\editor{}

\maketitle

\begin{abstract}

This paper { provides a theoretical justification of the superior classification performance} of deep rectifier networks over shallow rectifier networks from the geometrical perspective of piecewise linear (PWL) classifier boundaries. We show that, for a given threshold on the approximation error, the required number of boundary facets to approximate a general smooth boundary grows exponentially with the dimension of the data, and thus the number of boundary facets, { referred to as} {\it boundary resolution}, of a PWL classifier is an important quality measure that can be used to estimate a lower bound on the classification errors.   However, learning { naively} an exponentially large number of boundary facets requires the determination of an exponentially large number of parameters and also requires an exponentially large number of training  patterns. To overcome this issue of ``curse of dimensionality", compressive representations of high resolution classifier boundaries are required. To show the superior compressive power of deep rectifier networks over shallow rectifier networks, we prove that the maximum boundary resolution of a single hidden layer rectifier network classifier grows exponentially with the number of units when this number is smaller than the dimension of the patterns. When the number of units is larger than the dimension of the patterns, the growth rate is reduced to a polynomial order. Consequently, the capacity of generating a high resolution boundary will increase if the same large number of units are arranged in multiple layers instead of a single hidden layer. Taking  high dimensional spherical boundaries  as examples, we show how deep rectifier networks can utilize geometric symmetries to approximate a boundary  with the same accuracy but with a significantly fewer number of parameters than single hidden layer nets.

\end{abstract}


\section{Introduction}

Although as few as one hidden layer neural networks are capable of approximating any Borel measurable functions \citep{hornik1989multilayer}, deeper neural networks have outperformed  shallower neural networks in a wide range of applications such as  handwritten digit recognition \citep{ciresan2012multi}, object detection \cite{ren2015faster} and image classification \citep{krizhevsky2012imagenet,he2016deep,he2016identity}. The benefits of { neural networks' depth} have been investigated extensively in recent years, from the superior power of deep networks in function approximation \citep{delalleau2011shallow, eldan2016power, cohen2015expressive, mhaskar2016learning}, to the superior capacity of deep networks in separating the input space into a large number of regions of linearity \citep{Pascanu+et+al-ICLR2014b, montufar2014number, raghu2016expressive}. From these theoretical analyses, one can conclude that, for some functions that can be represented or approximated by both deep neural networks and single hidden layer networks, the representation { provided} by deep networks can be much more compact (i.e, with fewer parameters) and thus generalises better. However, { the functions considered so far have been} limited to certain families of polynomial functions or hand-coded functions, { which have been chosen} to demonstrate the expressive power of deep neural networks. It is unlikely that practically trained neural networks will fall into one of these analysed categories of functions. A good understanding of how and why deep neural networks achieve their empirical successes is thus still missing \citep{raghu2016expressive}.    

This paper aims to theoretically justify { the superior classification performances} of deep rectifier networks over shallow rectifier networks from the geometrical perspective of PWL classifier boundaries.  Given a dataset { comprising}  several classes, according to learning theory \citep{vapnik1998statistical}, a simpler learning model with a smaller sample complexity usually generalises better. However,  { approximation models} of a simple geometric surface can be quite complex, and a large { difference} may exist between the complexity of the approximation model and that of the original model. For instance,  \citep{dudley1974metric} shows that an exponentially large number of facets is required to approximate a spherical surface in $d$ dimensional space. This is despite the fact that a spherical surface is a simple geometric model which can be { represented} with $d+1$ parameters (one for the radius, the others for the center). We will show that the required number of units is at least a polynomial function of the dimension of the patterns if single hidden layer rectifier networks are used to approximate a spherical boundary (Theorem 7, Sec. 5). We will present an upper bound on the number of facets with respect to the number of units of a single hidden layer rectifier network. This upper bound shows that the capacity of single hidden layer rectifier nets { to generate} facets  increases exponentially when the number of units is smaller than the dimension of the input. However, the growth rate is reduced to a polynomial order when the number of units is larger than the dimension of the patterns (Lemma 6, Sec. 5). Consequently, the capacity of generating facets will increase if the same large number of units are arranged in multiple layers instead of a single hidden layer. With spherical surfaces as examples,  we will show that deep rectifier nets can be exponentially more efficient than single hidden layer nets. { The main contributions of this paper include: 
\begin{enumerate}
\item[i)] {\bf The introduction of boundary resolution for PWL classifiers} (Sec. 3). The resolution of PWL classifier boundaries is a measure of the classifier quality, which can be used to estimate a lower bound on the classification errors.  The introduction of this new concept provides a new { approach} to analyse the benefits of rectifier networks' depth. 
\item[ii)] {\bf The first investigation on the efficiency of deep rectifier networks in approximating  class boundaries} (Sec. 5-6). Given that the ultimate goal of deep learning for classification tasks is to learn class boundaries { rather than} classifier functions, it is critical to investigate the properties of deep neural networks in generating class boundaries in order to understand the benefits of networks' depth for classification tasks.  
\item[iii)] {\bf An explicit upper bound is provided on the number of facets that a single hidden layer rectifier can generate with a given number of units} (Sec. 5). This upper bound shows the limitations of single hidden layer networks and implies that deep nets have the potential to improve efficiency. For approximations of general convex boundaries  in $\mathbb{R}^d$, we show that the required number of facets is $O\left(\left(\frac{d}{\epsilon}\right)^{\frac{d-1}{2}}\right)$ for a threshold $\epsilon$ on the approximation error. To learn a convex PWL classifier with { this} large number of facets using a simple maxout network of some linear units,  $O\left(\left(\frac{d}{\epsilon}\right)^{\frac{d-1}{2}}\right)$ linear units are required and $O\left(d\left(\frac{d}{\epsilon}\right)^{\frac{d-1}{2}}\right)$ weights need to be learnt (Sec. 4). To use single hidden layer rectifier networks, the number of weights to be learnt is at least $O\left(d^2\left(\frac{d}{\epsilon}\right)^{\frac{1}{2}}\right)$  { for approximations of general convex boundaries (including spherical surfaces).}  (Sec. 5).
\item[iv)] {\bf The superior compressive power of deep rectifier networks is demonstrated by  { constructing a rectifier network} for spherical boundary approximation}.  The constructed network can learn the spherical boundary with $O\left(d\log\left(\frac{d}{\epsilon}\right)\right)$ units, each requiring at most 5 parameters to be learnt (Sec. 6).  As deep rectifier networks have the advantage to use a significantly smaller number of parameters to model the classifier boundaries, they usually thus generalize better than single hidden layer rectifier neural networks.

\end{enumerate}
}
  
 The rest of this paper is organised as follows. Section 2 addresses related work. Section 3 defines the resolution of PWL classifier boundaries and relates the resolution of convex PWL classifier boundaries to the number of linear units in the maxout representation of convex PWL classifiers. Section 4 addresses the required resolution of general convex classifier boundaries and show that an exponentially large number of facets is required to approximate a convex boundary even for spherical surfaces (the simplest convex surfaces). Section 5 presents the limit of single hidden layer rectifier networks in reducing the number of weights to be learnt for high resolution boundary representation. In Section 6, a solution with deep rectifier networks for the approximation of Euclidean balls is presented to show its superior efficiency over single hidden layer nets. Concluding remarks are provided in Section 7.

\section{Related Work}

The depth of neural networks has been investigated extensively in recent years to show the superior expressive power of deep neural networks over shallow networks.  \citet{delalleau2011shallow} showed that the deep network representation of a certain family of polynomials can be much more compact (i.e., with fewer hidden units) than that provided by a shallow network. Similarly, with the same number of hidden units, deep networks are able to split their input space into many more regions of linearity than their shallow counterparts \citep{Pascanu+et+al-ICLR2014b,montufar2014number}. 
\citet{eldan2016power} presented  an example function that is expressible by a small 3-layer neural networks, but cannot be approximated by a 2-layer network to a certain constant accuracy unless its width is exponential to the dimension of the data. \citet{cohen2015expressive} proved that, except for a negligible set, all functions that can be implemented by a deep network of polynomial size, require an exponential size in order to be realized (or even approximated) by a shallow network. 
\citet{mhaskar2016learning} demonstrated that deep networks can approximate the class of compositional functions with the same accuracy as shallow networks, but with an exponentially lower number of training parameters. The superior expressive power of deep residual nets
was analysed in \citep{veit2016residual} which showed that residual nets can be viewed as a collection of many paths of differing lengths, enabling very deep networks by activating only the short paths during training.

The most related work to this paper is the analysis about { the maximum number of regions that can be split by a rectifier neural network with a given number of units}. With hand-coded  construction of deep rectifier nets, \citet{montufar2014number} showed that deep nets can be exponentially more efficient in splitting the space into a large number of regions, while \citet{raghu2016expressive} presented an upper bound on the number of the split regions by a single hidden layer network with a given number of linear units. From these results, one can conclude that the complexity on the split regions by rectifier networks could grow exponentially with depth. However, it is not yet understood why the growth in complexity on the split regions of linearity improves generalization performance.  

All these related works focus on the properties of functions that can be represented by neural networks. However, the ultimate goal of classification is to find class boundaries whose function representations, however, are not unique.  Furthermore, the complexity of the different function representations of one identical class boundary  can be arbitrarily large. { For example, the boundary $\{\mbx: g\{f(\mbx)\}=0\}$ is the same as the boundary $\{\mbx:f(\mbx)=0\}$ for any strictly increasing function $g(z)$ with $g(0)=0$}. Consequently, the complexity analysis on the functions represented by neural networks is not directly on the complexity of class boundaries. Fortunately, there  is a rich history on the approximations of Euclidean Balls \citep{gordon1994volume} and general convex bodies \citep{macbeath1951extremal}, which will be shown to be closely related to rectifier/maxout networks. This paper will investigate the approximations of convex class boundaries and show the superior power of deep rectifier networks over single hidden layer networks.

\section{Resolution of PWL Classifier Boundaries}

In this section, we first define the resolution of a PWL classifier boundary as the number of exposed facets on the boundary, and consider the resolution of convex PWL classifier boundaries in particular. Since a general class boundary consists of one or more convex or concave subsets, the required resolution for the approximation of convex boundaries provides a lower bound on the required resolution for the approximation of general class boundaries.

Consider the boundary, namely 
\begin{equation}
\mathcal{B}_f\triangleq\{\mbx : f(\mbx)=0\},
\end{equation}
of a binary PWL classifier  $f(\mbx)$ which classifies the patterns $\mbx$ to be positive if $f(\mbx)>0$ or negative otherwise.
Since $f(\mbx)$ is a PWL classifier, its boundary consists of a number of facets each satisfying a linear equation $\mba_k^T\mbx+c_k=0$, more precisely
\begin{equation}
\begin{array}{rcl}
\mathcal{B}_f&=&\displaystyle\bigcup_{k=1}^n \Omega_k\\
\Omega_k&\triangleq& \{\mbx : \mba_k^T\mbx+b_k=0, f(\mbx)=0\}.
\end{array}
\end{equation}
Without loss of generality, we assume that none of the facets is redundant, that is,   
\begin{equation}
\bigcup_{k\not= i} \Omega_k \not= \mathcal{B}_f,\;\forall\; i.
\end{equation}
If there are any redundancies, one can always remove them and the remaining is then a set of facets without redundancy.  We say $\mathcal{B}_f$ has $n$ facets if it has $n$ distinct facets and  none of them is redundant.

Since any continuous function can be approximated by a PWL function, the boundary of any classifier that can be represented by a continuous function can also be approximated by a sufficiently larger number of facets around the boundary. For high accuracy approximation, a large number of facets is required in general. The resolution of PWL classifier boundaries is similar to the resolution of digital images, the later represents the quality of  digital images for the approximation of natural scenes while the former represents the quality of PWL classifiers for the approximation of class boundaries. In this paper, we focus on the approximations of convex boundaries with convex PWL classifiers. A general smooth class boundary can be viewed as the union of a number of convex/concave surfaces. Next, we address the resolution of convex PWL classifier boundaries.   

\subsection{Convex PWL classifiers and Their Resolutions}

A PWL function $f(\mbx)$ is said to be convex if the set $\{\mbx: f(\mbx)\leq 0\}$ is convex. For a convex PWL function $f(\mbx)$ in $\mathbb{R}^d$, the set $\{\mbx: f(\mbx)\leq 0\}$ is a polytope which can be described as the intersection of a finite number, say $m$, of half planes, i.e., 
\begin{equation}
\{\mbx: f(\mbx)\leq 0\} = \bigcap_{i=1}^m \{\mbx:\mbw_i^T\mbx+b_i\leq 0\}=\{\mbx:\max_{1\leq i\leq m}\mbw_i^T\mbx+b_i\leq 0\}
\end{equation}    
where $\mbw_i\in\mathbb{R}^d$ and $b_i\in\mathbb{R}$.   Therefore, any convex PWL classifier has a maxout representation, i.e.
\begin{equation}
f(\mbx) = \max_{1\leq i\leq m}\mbw_i^T\mbx+b_i.
\label{mxOut}
\end{equation}
A maxout representation, as in Eq. (\ref{mxOut}), of a convex PWL classifier $f(\mbx)$ is said to be irreducible if 
\begin{equation}
\{\mbx: f(\mbx)\leq 0\}\not\subset \{\mbx: w_k^T\mbx+b_k\leq 0\}, \;\forall\; 1\leq k \leq m.
\end{equation}

A convex PWL classifier with an irreducible maxout representation of $m$ units has $m$ facets on its boundary $\mathcal{B}_f$ and therefore its resolution is exactly the number of units in its maxout representation. Hereafter, we assume that the maxout representation of a PWL classifier is irreducible. If any unit is reducible, one can always remove it and the classifier remains the same (i.e., the classified label of any pattern $\mbx$ is invariant). For convenience, we use $\mathcal{M}_m$ to denote the set of convex PWL classifiers with irreducible maxout representation of $m$  units, i,e,
\begin{equation}
\mathcal{M}_m\triangleq \{f(\mbx)=\max_{1\leq i\leq m}\mbw_i^T\mbx+b_i: \mbw_i\in\mathbb{R}^d,b_i\in\mathbb{R}\}. 
\label{Mm}
\end{equation}

For any convex PWL classifier $f(\mbx)\in\mathcal{M}_m$, we will use $P_f$ to denote the polytope where $f(\mbx)$ is not positive, i.e.,  
\begin{equation}
P_f\triangleq \{\mbx:f(\mbx)\leq 0\}.
\end{equation}

In the next section, we will consider the required resolutions of convex PWL classifiers for the approximation of general convex boundaries. The required resolution (i.e., the number of required facets) will be used in Section 5 to estimate the number of required units for a single hidden layer rectifier network to approximate general smooth convex boundaries in high dimensional spaces.  

\section{Required Resolution for the Approximations of Smooth Convex Boundaries}

This section starts with spherical boundaries and then moves to general smooth convex boundaries. Surprisingly, for a given threshold on the approximation errors, the required resolution for a general convex surfaces is no higher than the required resolution for spherical surfaces (the simplest convex surfaces except for hyperplanes).  

\subsection{Approximation of Spherical Boundaries}  

{ Given that} the boundary of a convex body is convex and the convex combinations of the points in a convex boundary form a  convex body, the approximation of convex boundaries with convex PWL classifiers, which generates polytopic boundaries, is equivalent to the volume approximation of convex bodies with polytopes.  Next, we will use the results  on the polytopic approximations of Euclidean balls \citep{gordon1994volume} to derive the required number of facets to approximate spherical boundaries with given error thresholds. The approximation of general convex boundaries will be addressed in Section 4.2 using the results  on the polytopic approximations of general convex bodies \citep{macbeath1951extremal, schneider1967allgemeine,paouris2017random}.  
From \citep{gordon1994volume} [Theorem 5.2], we have 

\begin{prop} Let $B_2^d$ be the unit Euclidean ball in $\mathbb{R}^d$:
\begin{equation}
B_2^d=\left\{\mbx\in\mathbb{R}^d : \Vert \mbx\Vert^2\leq 1\right\}.
\end{equation}
There exists two constants $C_1, C$ (i.e., independent of $d$) such that for every integer $n\geq C_1 \left(\log d\right)^d$, it is possible to construct a polytope $P_n$ containing $B_{2}^d$, with at most $n$ facets, such that
\begin{equation}
\frac{\vert P_n\backslash B_2^d\vert}{\vert B_2^d\vert}\leq C\frac{d}{n^{\frac{2}{d-1}}} 
\end{equation}
where the $\vert S \vert$ represents the volume of a compact set $S$ and 
\begin{equation}
P_n\backslash B_2^d \triangleq \left\{\mbx: \mbx\in P_n, \mbx\not\in B_2^d\right\}. 
\end{equation}  
\end{prop} 

Proposition 1 shows that, for a given threshold $\epsilon$ on the approximation error, the number of facets required to approximate an Euclidean ball is 
\begin{equation}
\left(\frac{Cd}{\epsilon}\right)^{\frac{d-1}{2}}. 
\end{equation}

Note that
\begin{equation}
O\left(\left(\log d\right)^d\right) \ll O\left(\left(\frac{d}{\epsilon}\right)^{\frac{d-1}{2}}\right),
\end{equation}
we have the following corollary from Proposition 1:
\begin{coro} 
 For the unit Euclidean ball $B_2^d$ in $\mathbb{R}^d$, there exists a constant $C$(i.e., independent of $d$) such that for every small positive number $\epsilon$ and every integer
\begin{equation}
n\geq \left(\frac{Cd}{\epsilon}\right)^{\frac{d-1}{2}},
\end{equation}
it is possible to construct a polytope $P_n$ containing $B_{2}^d$, with at most $n$ facets, such that
\begin{equation}
\frac{\vert P_n\backslash B_2^d\vert}{\vert B_2^d\vert}\leq \epsilon. 
\end{equation} 
\end{coro}

Corollary 2 provides the required number of facets to approximate Euclidean balls. The next section will generalize this result to general convex bodies and show that the required resolution is no higher than the approximation of Eulidean balls.

\subsection{Approximation of General Convex Boundaries}

Let $\mathcal{P}_m$ be the set of convex polytopes with $m$ vertices, $K$ be a convex body (i.e., a bounded closed convex set with inner points), both in $\mathbb{R}^d$. Then define the function  of a convex body
\begin{equation}
\phi_m(\mathcal{K}) \triangleq \sup_{P\subset K, P\in \mathcal{P}_m} \frac{\vert P\vert}{\vert K\vert}
\end{equation}
where $\vert \cdot\vert$ denotes the volume of a convex body. According to \cite{macbeath1951extremal},
\begin{equation}
\phi_m(K)\geq \phi_m(B_2^d)
\end{equation}
holds for any convex body $K$.

Similarly, let $\hat{\mathcal{P}}_m$ be the set of convex polytopes with $m$ facets and define the function of a convex body
\begin{equation}
\hat{\phi}_m(\mathcal{K})\triangleq \inf_{K\subset P, P\in \hat{\mathcal{P}}_m} \frac{\vert P\vert}{\vert K\vert}.
\end{equation}
From \citep{schneider1967allgemeine} (or \citep{paouris2017random}[Corollary 5.2] for a recent reference), we have  
\begin{equation}
\hat{\phi}_m(K)\leq \hat{\phi}_m(B_2^d).
\end{equation} 

This indicates that, for a given threshold on the approximation errors, the approximation of a general convex body requires a smaller or equal number of facets than the approximation of the Euclidean balls. Combine the above discussions with Corollary 2 and the definition of $\mathcal{M}_m$ in Eq. (\ref{Mm}), we have  

\begin{thm}
Let $\epsilon>0$ be a small positive number. There is a constant $C$ (independent of the dimension $d$) such that for any bounded convex body $\mathcal{K}$ and any
\begin{equation}
m\geq \left(\frac{Cd}{\epsilon}\right)^{\frac{d-1}{2}}, 
\end{equation}
there exists $f(\mbx)\in \mathcal{M}_m$ such that $\mathcal{K}\subset P_f$ and
\begin{equation}
\frac{\vert P_f\backslash \mathcal{K}\vert}{\vert \mathcal{K}\vert}\leq \epsilon
\end{equation}
where $P_f\triangleq\{\mbx : f(\mbx) \leq 0\}$ is a polytope with $m$ facets.
\end{thm}

Theorem 3 shows that the representation of a general convex boundary requires an exponentially high resolution no matter the shape of the boundary is simple or complex. Learning the parameters of the linear units in a maxout network to approximate a spherical boundary will be more prone to overfitting than searching for the center and radius directly. To overcome the overfitting problem, more compact PWL  representations are required. Thanks to the symmetry of Euclidean balls, a group of facets that can be represented compactly with a small number of independent parameters, such as those in a regular polytope,  can be used to approximate a spherical boundary. In the next section, we will show  the limitations of single hidden layer nets in compressing the representation of high resolution class boundaries.  The superior compressive power of deep rectifier networks will be considered in Section 6.

 \section{Limit on the Compressive Power of Single Hidden Layer Rectifier Networks}
 
The number of facets that can be generated by a single hidden layer rectifier network is closely related to the number of regions into which the space is partitioned by the linear units of the network. In this section, we first give a brief review on results relating to the  partition of space $\mathbb{R}^d$ by Hyperplanes. We then show how these results relate to the compressive power of single hidden layer  rectifier networks.

Let $\mathcal{H}$ be a family of $m$ distinct hyperplanes in the $d$-dimensional space, we denote by $G_{\mathcal{H}}(d,m)$ the number of regions into which the space $\mathbb{R}^d$ is partitioned by the hyperplanes in $\mathcal{H}$. We denote by $G(d,m)$ the maximum of $G_{\mathcal{H}}(d,m)$ across all the possible sets, namely $\mathcal{H}$, of $m$ hyperplanes in $\mathbb{R}^d$, that is,
\begin{equation}
G(d,m)=\max_{\mathcal{H}} G_{\mathcal{H}}(d,m).
\end{equation}

A set $\mathcal{H}$ of $m$ hyperplanes, namely $\{\mbx: \mbw_i^T\mbx+b_i=0\}$, in $\mathbb{R}^d$ is said to be in {\it general position} if for each $1\leq k<d$, no $k+1$ members of $\mathcal{H}$ contain a common $(d-k)$-dimensional affine subset of $\mathbb{R}^d$, that is, any $(d+1)$ of $m$ hyperplanes has no common point if $m>d$, and all the $\mbw_i$  are independent if $m\leq d$. 

A region $D$ in $\mathbb{R}^d$ is called cone-like if whenever an $(d-1)$-dimensional plane cuts into components $D_1$ and $D_2$, such  that the cross-section is bounded, one of the two components will be bounded and the other unbounded.

From \cite{ho2006number}, we have 
\begin{prop}
The maximum number, denoted by $G(d,m)$, of regions in a $d$-dimensional space cut by $m$ hyperplanes is achieved when these $m$ hyperplanes are in general position and
\begin{equation}
G(d,m)=\dsum_{k=0}^d\left(\begin{array}{c}m\\k\end{array}\right).
\label{Gdm}
\end{equation}   
Furthermore, for any set of $m$ planes in general position in $\mathbb{R}^d$, among the $G(d,m)$ regions cut by these $m$ planes, 
$\left(\begin{array}{c}m-1\\d\end{array}\right)$ of them are bounded, and  each of the remaining 
\begin{equation}
C(d,m)\triangleq G(d,m)-\left(\begin{array}{c}m-1\\d\end{array}\right) = 2\dsum_{k=0}^{d-1}\left(\begin{array}{c}m-1\\k\end{array}\right)
\label{Cdm}
\end{equation}
unbounded regions is always cone-like.
\label{SHLbounds}
\end{prop}

Next, we present the main result of this section, regarding the limit on the resolution of single hidden layer rectifier networks. 

\begin{thm}
Let $\mathcal{SHL}(d,m)$ be the set of PWL functions that can be described by a single hidden layer rectifier network with $m$ rectifier linear units in $\mathbb{R}^d$, i.e.,
\begin{equation}
\mathcal{SHL}(d,m)=\left\{f(\mbx)=\mba^T\max(0,W\mbx+\mbb)+c: \mba,\mbb\in\mathbb{R}^m, W\in\mathbb{R}^{m\times d},c\in\mathbb{R},\mbx\in\mathbb{R}^d\right\}.
\end{equation} 
where $\mba,W,\mbb, c$ are parameters of the function while $\mbx$ is the variable.
Then the maximum number of facets of the boundaries $\mathcal{B}_f (\triangleq \{\mbx : f(\mbx)=0\}) $ across all the PWL functions in $\mathcal{SHL}(d,m)$ is between $C(d,m)$ and $G(d,m)$, that is 
\begin{equation}
C(d,m)-1\leq \max_{f(\mbx)\in \mathcal{SHL}(d,m)} \# (\mathcal{B}_f) \leq G(d,m)
\end{equation}
where $\#(\mathcal{B}_f)$ denotes the number of facets of $\mathcal{B}_f $, and $G(d,m),C(d,m)$ are defined in  Eq. (\ref{Gdm}) and Eq. (\ref{Cdm}) respectively. 
\end{thm}

{\bf Proof}: { Let $h_i(\mbx)=\mbw_i^T\mbx+b_i$ ($1\leq i\leq m$) be the $m$ linear units of $f(\mbx)=\mba^T\max(0,W\mbx+\mbb)+c$ where $\mbw_i^T$ is the $i^{th}$ row of $W$ and $b_i$ is the $i^{th}$ element of $\mbb$.  According to Proposition \ref{SHLbounds}, the space $\mathbb{R}^d$ can be cut into at most $G(d,m)$ regions by the $m$ hyperplanes $\{\mbx: h_i(\mbx)=0\}$. Within each of these regions, the sign of $h_i(\mbx)$ is invariant and thus $f(\mbx)$ is a linear function when $\mbx$ is constrained { within} one of these regions. Therefore the boundary $\mathcal{B}_f$ has at most one facet in each of these regions  and the maximum number of facets of $\mathcal{B}_f$ across all the functions in $\mathcal{SHL}(d,m)$ is upper bounded by $G(d,m)$.

For the lower bound $(C(d,m)-1)$, we consider the conditions under which $\mathcal{B}_f$ has exactly $(C(d,m)-1)$ facets. First, let us choose the linear units such that the associated $m$ hyperplanes $\{h_i(\mbx)=0\}$ are in {\it general } position. Then, according to Proposition \ref{SHLbounds}, these $m$ hyperplanes cut the space $\mathbb{R}^d$ into $G(d,m)$ regions where $C(d,m)$ of them are unbounded. Since $f(\mbx)$ is linear when $\mbx$ is constrained { within} each of the
$G(d,m)$ regions, $f(\mbx)$ has a bounded maximum in the $(G(d,m)-C(d,m))$ bounded and closed regions. Furthermore, in the $C(d,m)$ unbounded regions, there is at most one region within which all the linear units $h_i(\mbx)$ are negative and therefore $f(\mbx)$ is unbounded in at least $(C(d,m)-1)$ of the unbounded regions. Hence, if all the elements of $\mathbf{a}^T$ are positive, and $c$ is sufficiently small (e.g. approaching to $-\infty$) such that $f(\mbx)$ is negative in all the bounded regions, then $\mathcal{B}_f$ has a facet in each of the $C(d,m)$ unbounded regions except the one (if any) within which all the linear units $h_i(\mbx)$ are negative. Thus, the maximum number of facets of $\mathcal{B}_f$ across all $f(\mbx)\in \mathcal{SHL}(d,m)$ is lower bounded by $(C(d,m)-1)$.   

}

\begin{flushright}
 $\blacksquare$
\end{flushright}

To estimate the required number of units to approximate a general convex boundary by an SHL rectifier network, the following lemma is also required.
\begin{lem} Let $G(d,m)$ be the maximum number  of regions in a $d$-dimensional space cut by $m$ hyperplanes, then
\begin{equation}
G(d,m)=2^m, \;\forall\; m\leq d,
\label{newIneq}
\end{equation}
\begin{equation}
G(d,m) \leq \frac{1}{(d-1)!}m^d,\;\forall\; m\geq d.
\label{Gineq}
\end{equation}
\end{lem}

{\bf Proof}: Note that
\begin{equation}
\dsum_{k=0}^m\left(\begin{array}{c}m\\k\end{array}\right)=2^m
\end{equation}
and $\left(\begin{array}{c}m\\k\end{array}\right)=0$ if $k>m$. Then Eq. (\ref{newIneq}) follows from Proposition 4. 

{
For the proof of Eq. (\ref{Gineq}), by induction, it suffices to prove the following three statements:
\begin{enumerate}
\item[i)] Eq. (\ref{Gineq}) holds when $d=2$;
\item[ii)] Eq. (\ref{Gineq}) holds when $m=d$;
\item[iii)] For any $k\geq 2, l\geq k+1$, if Eq. (\ref{Gineq}) holds when $m=l, d=k, k+1$, then  Eq. (\ref{Gineq}) holds when $m=l+1,d=k+1$. 
\end{enumerate} 

For the proof of statement i), it is easy to check that  $G(2,m)=m+1+\frac{m(m-1)}{2}$ and therefore $G(2,m)\leq m^2$ when $m\geq 2$. Thus Eq. (\ref{Gineq}) holds when $d=2$. This proves i).

Next we prove statement ii), i.e. 
\begin{equation}
2^d\leq \frac{1}{(d-1)!}d^{d}. 
\label{Ineq1}
\end{equation}
This can be done by induction. It is easy to check that Eq. (\ref{Ineq1}) holds when $d=2$. Now assume that Eq. (\ref{Ineq1}) holds when $d=k$ for some $k\geq 2$, that is 
\begin{equation}
2^k\leq \frac{1}{(k-1)!}k^{k}. 
\label{Ineq1k}
\end{equation}
Note that, for any integer $n\geq 2$ and any positive number $x>0$, we have 
\begin{equation}
(x+1)^n = \sum_{i=0}^n \left(\begin{array}{c}n\\ i
\end{array}\right) x^i\geq x^n+nx^{n-1}
\end{equation}
and therefore 
\begin{equation}
(k+1)^{k+1}\geq k^{k+1}+(k+1)k^k
\end{equation}
 which implies that
 \begin{equation}
\frac{1}{k!}(k+1)^{k+1}\geq \frac{k^{k+1}}{k!}+\frac{(k+1)k^k}{k!}\geq 2\frac{k^k}{(k-1)!}. 
 \end{equation}
Then, from (\ref{Ineq1k}), it follows 
 \begin{equation}
\frac{1}{k!}(k+1)^{k+1}\geq 2^{k+1}. 
 \end{equation}
 That is, Eq. (\ref{Ineq1}) holds for $d=k+1$ as well. This completes the induction and completes the proof of Eq. (\ref{Ineq1}).
 
For the proof of statement iii), assume that Eq. (\ref{Gineq}) holds when $m=l, d=k,k+1$, that is 
\begin{equation}
\begin{array}{rcl}
G(k,l)&\leq& \frac{1}{(k-1)!}l^{k}\\
G(k+1,l)&\leq& \frac{1}{k!}l^{k+1}.\\
\end{array}
\label{Ineq2}
\end{equation}
From the following recursive relations of combinations 
\begin{equation}
\left(\begin{array}{c}l+1\\ i
\end{array}\right)=\left(\begin{array}{c}l\\ i
\end{array}\right)+\left(\begin{array}{c}l\\ i-1
\end{array}\right),
\end{equation}
we have 
\begin{equation}
\begin{array}{rcl}
G(k+1,l+1)&=& \dsum_{i=0}^{k+1} \left(\begin{array}{c}l+1\\ i
\end{array}\right)\\
&=&\dsum_{i=0}^{k+1} \left(\begin{array}{c}l\\ i
\end{array}\right)+\dsum_{i=1}^{k+1} \left(\begin{array}{c}l\\ i-1
\end{array}\right)\\
&=&G(k+1,l)+\dsum_{j=0}^{k} \left(\begin{array}{c}l\\ j
\end{array}\right)\\
&=&G(k+1,l)+G(k,l).\\
\end{array}
\end{equation}
Then, from (\ref{Ineq2}), it follows that
\begin{equation}
\begin{array}{rcl}
G(k+1,l+1)&\leq& \frac{1}{k!}l^{k+1}+\frac{1}{(k-1)!}l^{k}\\
&=&\frac{1}{k!}\{l^{k+1}+kl^k\}\\[0.1cm]
&\leq&\frac{1}{k!}(l+1)^{k+1}
\end{array}
\label{Ineq3}
\end{equation}
which implies that Eq. (\ref{Gineq}) holds for $d=k+1,m=l+1$ and the proof  is completed. 
}

\begin{flushright}
 $\blacksquare$
\end{flushright}

Now we are ready to estimate the required number of units for a single hidden layer network to approximate general convex boundaries.  

From Eq. (\ref{Gineq}), to generate 
\begin{equation}
\left(\frac{Cd}{\epsilon}\right)^{\frac{d-1}{2}}
\end{equation}
 number of facets by a single hidden layer net, the number of units, namely $m$, should satisfy
\begin{equation}
\frac{1}{(d-1)!} m^{d}\geq \left(\frac{Cd}{\epsilon}\right)^{\frac{d-1}{2}},
\end{equation}
and therefore 
\begin{equation}
\begin{array}{rcl}
m&\geq& \left(\frac{Cd}{\epsilon}\right)^{\frac{d-1}{2d}} \left\{(d-1)!\right\}^{\frac{1}{d}}\\
&\approx& \left(\frac{Cd}{\epsilon}\right)^{\frac{1}{2}} \left\{(d-1)!\right\}^{\frac{1}{d}}
\end{array}
\end{equation}

From Stirling's formula for approximation of $n!$ \citep{romik2000stirling}, we have 
\begin{equation}
d! \geq \sqrt{2\pi}d^{d+0.5}e^{-d}
\end{equation}
and therefore
\begin{equation}
\begin{array}{rcl}
m&\geq& \left(\frac{Cd}{\epsilon}\right)^{\frac{1}{2}} \left\{(d-1)!\right\}^{\frac{1}{d}}\\
&\approx& \left(\frac{Cd}{\epsilon}\right)^{\frac{1}{2}}\frac{d-1}{e}
\end{array}
\end{equation}
where $e\approx 2.7183 $ is the Euler number. 

Theorem 7 below summarises the result of this section.

\begin{thm}
 To approximate the Euclidean ball in $d$ dimensional space with $\left(\frac{Cd}{\epsilon}\right)^{\frac{d-1}{2}}$ facets by a single hidden layer rectifier net, at least 
\begin{equation}
N_{s}=\left(\frac{Cd}{\epsilon}\right)^{\frac{1}{2}}\frac{d-1}{e}
\end{equation} 
units are required, with 
\begin{equation}
N_{sp}=(d+1)N_s=\left(\frac{Cd}{\epsilon}\right)^{\frac{1}{2}}\frac{d^2-1}{e}
\end{equation} 
parameters to be learnt. 
\end{thm}

In the next section, we will show that a deep rectifier net is able to approximate Euclidean balls with a much smaller number of units and parameters.

\section{Superior Compressive Power of Deep Rectifier Networks} 
 
 We will first construct a deep rectifier network for two dimensional data in Section 6.1 and then construct a deep  rectifier network in Section 6.2 for efficient approximation of high dimensional spherical boundaries.

\subsection{Two Dimensional Space}

Consider the approximation of the unit circle $\{\mbx \in \mathbb{R}^2 : \Vert \mbx\Vert = 1\}$. From Theorem 5, if a single hidden layer net of $m$ units is used, the maximum number of segments is $G(2,m)$($=m+1+\frac{m(m+1)}{2}$). Next we present a deep rectifier net which can approximate the unit circle more efficiently than single hidden layer nets.

\begin{lem}
Let $\theta_k=\dfrac{\pi}{2^k}$, and  $f_k(x,y), \bar{f}_k(x,y)$ be defined recursively as  
\begin{equation}
\begin{array}{rcl}
f_1(x,y)&=&\vert x \vert,\bar{f}_1(x,y)=\vert y\vert\\
f_{k+1}(x,y)&=&\cos \theta_{k+1} f_k(x,y)+\sin\theta_{k+1} \bar{f}_k(x,y)\\
\bar{f}_{k+1}(x,y)&=&\left\vert -\sin\theta_{k+1}f_k(x,y)+\cos\theta_{k+1} \bar{f}_k(x,y)\right\vert.\\
\end{array}
\end{equation}
Then for any $[x,y]^T\in \mathbb{R}^2$, $k=1,2,\cdots,$ we have
\begin{equation}
\begin{array}{rcccl}
1-\frac{\pi^2}{2^{2k+1}}<\cos\theta_k &\leq& \frac{f_k(x,y)}{\sqrt{x^2+y^2}} &\leq& 1.
\end{array}
\end{equation}
\label{lem2D}
\end{lem}

{\bf Proof}: Let $f_0(x,y)=x=r \cos t, \bar{f}_0(x,y)=y=r\sin t, t\in [0,2\pi]$ where $r=\sqrt{x^2+y^2}$, then
\begin{equation}
f_1(x,y)= \vert x\vert =r \vert \cos t\vert= r\cos t_1; \bar{f}_1(x,y)=\vert y\vert=r\vert \sin t\vert=r\sin t_1 
\end{equation}
where  $t_1\in \left[0,\frac{\pi}{2}\right]$ and 
\begin{equation}
t_1=\left\{\begin{array}{lcl} t; \;&\mathrm{if}&\; t\in [0,\pi/2];\\
\pi-t; \; &\mathrm{if}&\; t\in (\pi/2,\pi];\\ 
t-\pi; \; &\mathrm{if}&\; t\in (\pi,3\pi/2];\\
2\pi -t\; &\mathrm{if}&\; t\in (3\pi/2,2\pi].
\end{array}\right.
\end{equation}

Next assume that $f_k=r\cos t_k, \bar{f}_k=r\cos t_k$ holds some $k\geq 1$ and some $t_k\in [0,\theta_k]$, where $\theta_k=\frac{\pi}{2^k}$. 
Then
\begin{equation}
\begin{array}{rcl}
f_{k+1}&=& r\cos\theta_{k+1}\cos t_k+r\sin\theta_{k+1}\sin t_k =r\cos (t_k-\theta_{k+1})\\
\bar{f}_{k+1}&=& r\left\vert -\sin\theta_{k+1}\cos t_k+\cos\theta_{k+1}\sin t_k\right\vert =r\vert \sin (t_k-\theta_{k+1})\vert.\\
\end{array}
\end{equation}
Let 
\begin{equation}
t_{k+1}=\vert t_k-\theta_{k+1}\vert\in [0,\theta_{k+1}],
\end{equation}
then $f_{k+1}=r\cos t_{k+1}, \bar{f}_{k+1}=r\sin t_{k+1}$. By induction, for any $k\geq 1$, $f_{k}=r\cos t_{k}, \bar{f}_{k}=r\cos t_{k}$ holds for some $t_k\in [0,\theta_k]$ and therefore 
 \begin{equation}
 \begin{array}{rcl}
 \frac{f_k}{\sqrt{x^2+y^2}}= \frac{r\cos t_k}{r}\geq \cos\theta_k
 \end{array}
 \end{equation}
which completes the proof. \hfill 
\begin{flushright}
$\blacksquare$
\end{flushright}

To implement the function $f_k(x,y)$ for some integer $k$, the rectifier net can be constructed as follows:
\begin{enumerate}
{\it
\item[1).] The first hidden layer has four nodes which are chosen as below: 
\begin{equation}
\begin{array}{rcl}
\mbz_1(1)&=& \max(x_1,0)\\
\mbz_1(2)&=& \max(-x_1,0)\\
\mbz_1(3)&=& \max(x_2,0)\\
\mbz_1(4)&=& \max(-x_2,0)\\
\end{array}
\end{equation}
\item[2).] The second hidden layer has three nodes:   
\begin{equation}
\begin{array}{rcl}
\mbz_2(1)&=&\max(0,\vert x_1\vert \cos\theta_2+\vert x_2\vert \sin\theta_2)\\
\mbz_2(2)&=&\max(0,\vert x_1\vert \cos\theta_2-\vert x_2\vert \sin\theta_2)\\
\mbz_2(3)&=&\max(0,\vert x_2\vert \sin\theta_2-\vert x_1\vert \cos\theta_2)\\
\end{array}
\end{equation}
where
\begin{equation}
\begin{array}{rcl}
\theta_2&=& \frac{\pi}{4}\\
\vert x_1\vert &=& \mbz_1(1)+\mbz_1(2)\\
\vert x_2\vert &=& \mbz_1(3)+\mbz_1(4).\\
\end{array}
\end{equation}
\item[3).] The $(i+1)^{th} (i=2,\cdots, k-2)$ layer has three nodes which are chosen as 
\begin{equation}
\begin{array}{l}
\mbz_{i+1}(1)=\max\left\{0,\mbz_i(1)\cos\theta_{i+1}+\left(\mbz_i(2)+\mbz_i(3)\right)\sin\theta_{i+1}\right\}\\
\mbz_{i+1}(2)=\max\left\{0,\mbz_i(1)\cos\theta_{i+1}+\left(\mbz_i(2)-\mbz_i(3)\right)\sin\theta_{i+1}\right\}\\
\mbz_{i+1}(3)=\max\left\{0,\left(\mbz_i(2)-\mbz_i(3)\right)\sin\theta_{i+1}-\mbz_i(1)\cos\theta_{i+1}\right\}\\
\end{array}
\end{equation}
where
\begin{equation}
\theta_{i+1}=\frac{\theta_{i}}{2}.
\end{equation}
\item[4).] The output layer is then
\begin{equation}
f_k=\mbz_{k-1}(1)\cos\theta_{k}+\left(\mbz_{k-1}(2)+\mbz_{k-1}(3)\right)\sin\theta_k-1
\end{equation}
where $\theta_k=\frac{\theta_{k-1}}{2}$.
}
\end{enumerate}

The constructed rectifier net has $3(k-1)+2=3k-1$ units in total and can generated $2^k$ segments to approximate a circle. The following Lemma summarises the main result of this section.
\begin{lem}
A rectifier net with $m=3k-1$ units in $k$ ($\geq 2$) layers exists to generate a polytope 
\begin{equation}
P_n=\{(x,y)^T : f_k(x,y)\leq 1\}, 
\end{equation}
 of $n=2^{\frac{m+1}{3}}$ number of segments, to approximate the unit circle $B_2^2$ such that $B_2^2\subset P_n$ and
\begin{equation}
\frac{\vert P_n-B_2^2\vert }{\vert B_2^2\vert}\leq \frac{1}{\cos^2\theta_k}-1\approx\theta_k^2=\frac{\pi^2}{2^{2k}}
\end{equation}
\end{lem}

{\bf Proof}: From Lemma 8, it follows that $B_2^2\subset P_n$ and 
\begin{equation}
x^2+y^2\leq \frac{1}{\cos^2\theta_r},\;\forall\; (x,y)\in P_n
\end{equation} 
which implies that $P_n\subset \{(x,y)^T: x^2+y^2\leq \frac{1}{\cos^2\theta_k}\}$ and therefore 
\begin{equation}
\frac{\vert P_n-B_2^2\vert }{\vert B_2^2\vert}\leq \frac{\vert\frac{1}{\cos\theta_k} B_2^2\vert-\vert B_2^2\vert }{\vert B_2^2\vert} = \frac{1}{\cos^2\theta_k}-1\approx\theta_k^2=\frac{\pi^2}{2^{2k}}.
\end{equation}
\begin{flushright}
 $\blacksquare$
\end{flushright}

\subsection{High Dimensional Spaces}

{ For high dimensional cases, we will  first use the constructed 2D rectifier network to approximate $\sqrt{x_1^2+x_2^2}$ (i.e., the norm of the first two elements). Then with $x_3$ as one input and the output of the previous 2D network as another input, we apply the same 2D network to approximate the norm of the first three elements. Sequentially, by applying the 2D rectifier net similarly for $(d-1)$ times, the norm of $d-$dimensional vectors can be approximated. The following lemma will give the bound for the approximation error and will be used to estimate the approximation error for the approximation of unit balls using the constructed deep rectifier network. 
}

\begin{lem}
Let $\mbx\in \mathbb{R}^d$ and $g_l(\mbx;k)$ be defined recursively as below
\begin{equation}
\begin{array}{rcl}
g_1(\mbx;k)&=& f_k(x_1,x_2)\\
g_{l}(\mbx;k)&=& f_k\{g_{l-1}(\mbx;k),x_{l+1}\}, 2\leq l\leq d-1\\
\end{array}
\end{equation}
 where $f_k(\cdot,\cdot)$ is a function defined in Lemma \ref{lem2D} with $\theta_k=\frac{\pi}{2^k}$. 
Then for any $\mbx \in \mathbb{R}^d$, $k=1,2,\cdots,$ we have
\begin{equation}
\begin{array}{rcccl}
\cos^{l} \theta_k &\leq& \frac{g_{l}(\mbx;k)}{\sqrt{\sum_{i=1}^{l+1}x_i^2}} &\leq& 1,\;\forall\; 1\leq l\leq d-1.
\label{NormApprox}
\end{array}
\end{equation}
\label{lemNd}
\end{lem}

{\bf Proof}: { From Lemma \ref{lem2D}, we have 
\begin{equation}
\begin{array}{rcccl}
\cos \theta_k &\leq& \frac{g_1(\mbx;k)}{\sqrt{x_1^2+x_2^2}} &\leq& 1\\
\cos \theta_k &\leq& \frac{g_{l}(\mbx;k)}{\sqrt{g_{l-1}^2+x_{l+1}^2}} &\leq& 1, \;\forall\; l\geq 2 .\\
\label{Ineq2nd}
\end{array}
\end{equation}
Hence Eq. (\ref{NormApprox}) holds when $l=1$. 

Next, assume that Eq. (\ref{NormApprox}) holds for $l=p$ for some $p\geq 1$, that is 
\begin{equation}
\begin{array}{rcccl}
\cos^{p} \theta_k &\leq& \frac{g_{p}(\mbx;k)}{\sqrt{\sum_{i=1}^{p+1}x_i^2}} &\leq& 1.
\end{array}
\end{equation}
Then from Eq. (\ref{Ineq2nd}), we have 
\begin{equation}
g_{p+1}^2(\mbx;k)\leq g_{p}^2(\mbx;k) + x_{p+2}^2\leq \sum_{i=1}^{p+2}x_i^2
\label{Ineq1LemnD}
\end{equation}
and 
\begin{equation}
\begin{array}{rcl}
g_{p+1}^2(\mbx;k)&\geq& \cos^2\theta_k \left\{g_{p}^2(\mbx;k) + x_{p+2}^2\right\}\\
&\geq& \cos^2\theta_k \left(\cos^{2p}\theta_k \sum_{i=1}^{p+1} x_i^2 + x_{p+2}^2\right)\\
&\geq& \cos^{2p+2}\sum_{i=1}^{p+2} x_i^2.
\end{array} 
\label{Ineq2lemnD}
\end{equation}
which, together with Eq. (\ref{Ineq1LemnD}), imply that Eq. (\ref{NormApprox}) holds for $l=p+1$. By induction, Eq. (\ref{NormApprox}) is true for every $l=1,2,\cdots,d-1$. 

\begin{flushright}
 $\blacksquare$
\end{flushright}

Next, we use Lemma 10 to estimate the required number of units to approximate Euclidean balls with a given threshold on the approximation errors. Consider the polytope 
\begin{equation}
P\triangleq \{\mbx: g_{d-1}(\mbx;k)\leq 1\}
\end{equation}
where $g_{d-1}(\mbx;k)$ is defined in Lemma 10. 

From Lemma 10, we have 
\begin{equation}
\Vert \mbx\Vert \cos^{d-1}\theta_k \leq g_{d-1}(\mbx;k)\leq \Vert \mbx\Vert
\end{equation}
and therefore
\begin{equation}
B_2^d \subset P\subset B\left(\frac{1}{\cos^{d-1}\theta_k}\right) 
\end{equation}
where $B(r)\triangleq \{\mbx: \Vert \mbx\Vert\leq r\}$ is an Euclidean ball in $\mathbb{R}^d$ with radius $r$. Note that the volume of an Euclidean ball in $\mathbb{R}^d$ with radius $r$ is proportional to $r^d$, the approximation error of $P$ to the unit ball $B_2^d$ satisfies  
\begin{equation}
\begin{array}{rcl}
\frac{\vert P\backslash B_2^d\vert}{\vert B_2^d\vert} &<& \left\{\frac{1}{\cos^{d-1}\theta_k}\right\}^d-1 \\
&=& \left\{\frac{1}{\cos\theta_k}\right\}^{(d-1)d}-1\\
\end{array}
\end{equation}
Note that $\theta_k=\frac{\pi}{2^k}$ is very close to zero when $k\geq 10$, and when  $x$ is close to zero,  
\begin{equation}
\begin{array}{rcl}
\left(\frac{1}{\cos x}\right) &\approx& 1 + \frac{1}{2}x^2\\
(1+\frac{1}{2}x^2)^n&\approx& 1+\frac{n}{2}x^2.
\end{array} 
\end{equation}
We have 
\begin{equation}
\begin{array}{rcl}
\frac{\vert P\backslash B_2^d\vert}{\vert B_2^d\vert} &<& \frac{d(d-1)}{2}\theta_k^2\\
&=& \frac{d(d-1)\pi^2}{2^{2k+1}}.
\end{array}
\end{equation}

}

To meet a threshold $\epsilon$ on the approximation error, it suffices to choose $k$ such that
\begin{equation}
2k+1\geq 2\log(d)+\log(\epsilon^{-1})+2\log(\pi)
\end{equation}
or  equivalently
\begin{equation}
k \geq \log(d)+\frac{1}{2}\log(\epsilon^{-1})-\frac{1}{2}+\log(\pi).
\end{equation}

The following Theorem summarises the result of this section.
\begin{thm}
Let
\begin{equation}
k^* = \log(d)+\frac{1}{2}\log(\epsilon^{-1})-\frac{1}{2}+\log(\pi).
\end{equation}
There exists a rectifier network, with $k^*(d-1)$ layers of 
\begin{equation}
N_d=(d-1)(3k^*-1)= (d-1)\{3\log(d)+1.5\log(\epsilon^{-1})+3\log(\pi)-2.5\}
\end{equation}
 units, which can approximate a $d$ dimensional Euclidean ball such that 
\begin{equation}
\frac{\vert P_n\backslash B_2^d\vert}{\vert B_2^d\vert}\leq \epsilon. 
\end{equation}
  
\end{thm}

\subsection{Advantages of Deep Rectifier Networks}

From Theorem 7 and Theorem 11, one can see that the ratio between $N_s$, the number of required units for single hidden layer nets to approximate Euclidean balls, and $N_d$ is 
\begin{equation}
\frac{N_s}{N_d}\approx \frac{\left(\frac{Cd}{\epsilon}\right)^{\frac{1}{2}}}{e\left(3\log d+1.5\log (\epsilon^{-1})\right)}
\end{equation} 
which shows that, with similar approximation accuracy, single hidden layer nets require much larger number of units than the constructed deep rectifier nets. In particular, when $d$ is small, the number of required facets is dominated by the approximation accuracy and the constructed deep net is exponentially (with the depth) more efficient than single hidden layer nets. Note that each node of the constructed deep rectifier network is connected to 3 or 4 other nodes, each unit has at most 5 parameters to be determined, while each unit of the single hidden layer net has $(d+1)$ parameters to learn. For large dimension $d$, the constructed deep rectifier network is at least  $O(d^{1.5}/\log(d))$ more efficient than single hidden layer neural networks.

\section{Concluding Remarks}

By introducing the boundary resolution of PWL classifiers, this paper has shown the superior compressive power of deep rectifier networks over single hidden layer rectifier networks for high resolution representation of class boundaries. Due to the requirement of the universal approximation capacity, non-polynomial activation functions such as rectifiers are used in neural networks, but at the cost of exponentially (with respect to data dimension) increased model complexity for the approximation of geometrically-simple class boundaries, such as spherical boundaries or other boundaries that can be represented by a small number of parameters. To learn such geometrically-simple boundaries, deep neural nets are required to learn compact models for the purpose of good generalization by exploiting the symmetric properties of class boundaries.

\begin{small}
\bibliography{icml2017bib}
\bibliographystyle{IEEEtran}
\end{small}


\end{document}